%% file: main.tex
\documentclass[
]{ceurart}

\usepackage{listings}
\begin{document}

\copyrightyear{2024}
\copyrightclause{Copyright for this paper by its authors.
  Use permitted under Creative Commons License Attribution 4.0
  International (CC BY 4.0).}

\conference{HI-AI@KDD, Human-Interpretable AI Workshop at the KDD 2024, 26$^{th}$ of August 2024, Barcelona, Spain }

\title{Ontology-grounded Automatic Knowledge Graph Construction by LLM under Wikidata schema}

\author[1]{Xiaohan Feng}[%
email=xhfeng@se.cuhk.edu.hk
]

\author[1]{Xixin Wu}[%
email=wuxx@se.cuhk.edu.hk
]

\author[1]{Helen Meng}[%
email=hmmeng@se.cuhk.edu.hk
]
\cormark[1]

\address[1]{Department of System Engineering and Engineering Management, Chinese University of Hong Kong}

\cortext[1]{Corresponding author.}

\begin{abstract}
We propose an ontology-grounded approach to Knowledge Graph (KG) construction using Large Language Models (LLMs) on a knowledge base. An ontology is authored by generating Competency Questions (CQ) on knowledge base to discover knowledge scope, extracting relations from CQs, and attempt to replace equivalent relations by their counterpart in Wikidata. To ensure consistency and interpretability in the resulting KG, we ground generation of KG with the authored ontology based on extracted relations.  Evaluation on benchmark datasets demonstrates competitive performance in knowledge graph construction task. Our work presents a promising direction for scalable KG construction pipeline with minimal human intervention, that yields high quality and human-interpretable KGs, which are interoperable with Wikidata semantics for potential knowledge base expansion.
\end{abstract}


\maketitle

\section{Introduction}
\input{src/intro}
\section{Literature Review}
\input{src/lit}
\section{Method: Ontology-grounded KG Construction}
\input{src/metho}
\section{Experiments and Discussion}
\input{src/res}
\section{Conclusion}
\input{src/con}


\begin{acknowledgments}
This work is supported by Centre for Perceptual and Interactive Intelligence (CPII) Ltd, a CUHK-led InnoCentre under InnoHK scheme of Innovation and Technology Commission. 
\end{acknowledgments}

\bibliography{bib}

\appendix
\input{src/app}

\end{document}

%% file: src/intro.tex
Knowledge Graphs (KGs) are structured representations of information that capture entities and their relationships in a graph format. By organizing knowledge in a machine-readable way, KGs enable a wide range of intelligent applications, such as semantic search, question answering, recommendation systems, and decision support \cite{10.1145/3447772}. The ability to construct high-quality, comprehensive KGs is thus critical for harnessing the power of these technologies across various domains.

Traditionally, the process of constructing KGs has relied heavily on manual effort by domain experts to define the relevant entities and relationships, populate the graph with valid facts, and ensure logical consistency \cite{Ji2020ASO}. However, this manual curation approach is time-consuming, expensive, and difficult to scale to large, evolving domains. There is a strong need for (semi-)automatic methods that can aid the KG construction process by extracting structured knowledge from unstructured data sources such as text.

Recent years have seen growing interest in leveraging Large Language Models (LLMs) for various knowledge capture and reasoning tasks \cite{zhu_llms_2024}. Pre-trained on vast amounts of text data, LLMs can generate fluent natural language and have been shown to memorize and recall factual knowledge \cite{Petroni2019LanguageMA}, \cite{openai2024gpt4}. 
However, directly applying LLMs to KG construction still faces several challenges. First, LLMs may generate inconsistent or redundant facts due to the lack of an explicit, unified schema \cite{agrawal_can_2024}. Second, the generated KGs may be incomplete or biased towards the knowledge present in the LLM's training data, which may not fully cover the target domain, especially for proprietary documents not included in pre-training set. Finally, it can be challenging to integrate LLM-generated KGs with existing knowledge bases due to misalignment with standard ontologies.


\input{src/flow}

In this work, we propose a novel approach that harnesses the reasoning power of LLMs and the structured schema of Wikidata to construct high-quality KGs for proprietary knowledge domains. Our approach begins by discovering the scope of knowledge through the generation of Competency Questions (CQ) and answers from unstructured documents. We then summarize the relations and properties from these QA pairs into an ontology, matching candidate properties against those defined in Wikidata and extending the schema as needed. Finally, we use the resulting ontology to ground the transformation of CQ-answer pairs into a structured KG. By incorporating the Wikidata schema into our pipeline and grounding generation of KG on the same ontology, we aim to reduce redundancy, leverage the implicit knowledge captured during LLM pretraining while improving interpretability, and ensure interoperability with public knowledge bases. The generated KGs could be parsed with RDF parsers and used in downstream applications, or audited for correctness.

The main contributions of this work are as follows:

\begin{enumerate}
    \item We propose a novel ontology-grounded approach to LLM-based KG construction that leverages ontology based on Wikidata schema to guide the extraction and integration of knowledge from unstructured text.
    \item We introduce a pipeline that combines competency question generation, ontology alignment, and KG grounding to systematically construct high-quality KGs that are consistent, complete, and interoperable with existing knowledge bases.
    \item We demonstrate the effectiveness of our approach through  experiments on benchmark datasets, showing improvements in KG quality compared to traditional methods alongside with interpretability and utility of generated KGs. 
\end{enumerate}

%% file: src/flow.tex
\begin{wrapfigure}{R}{0.5\textwidth}
    \centering
    \includegraphics[width=0.5\textwidth]{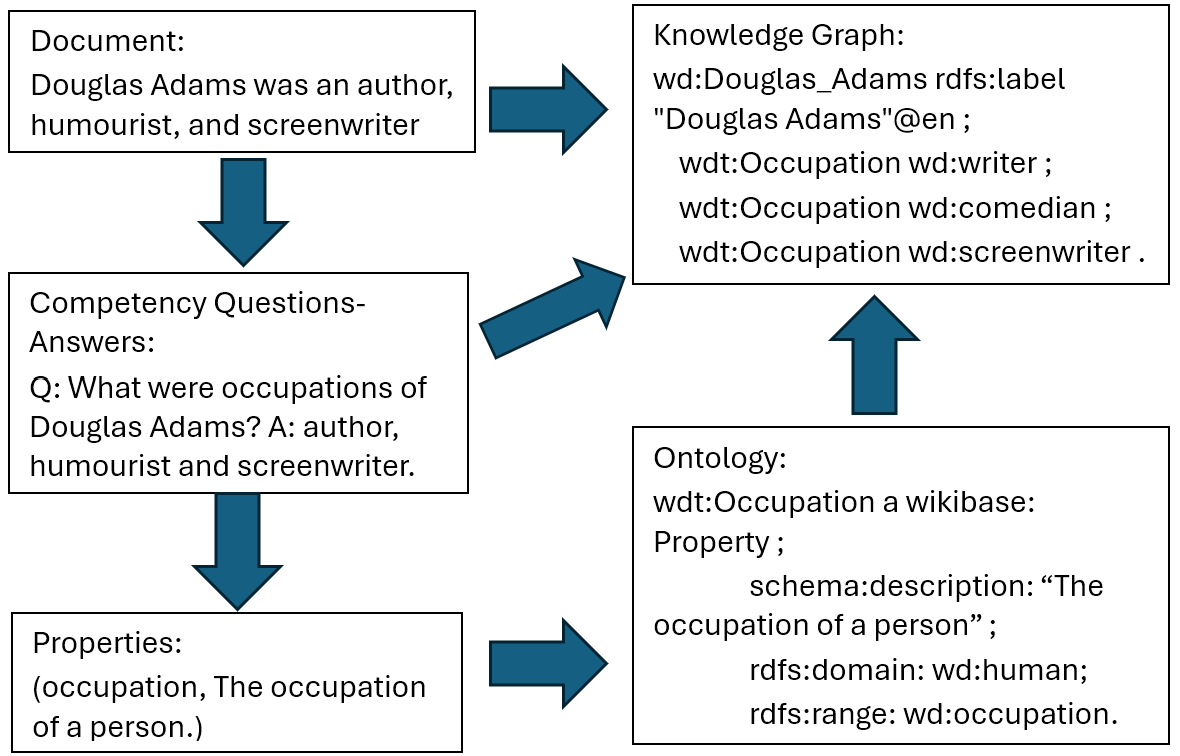}
    \caption{Flowchart of proposed approach.}
    \label{fig:figure}
\end{wrapfigure}

%% file: src/lit.tex
Knowledge graph construction has been an active area of research in recent years, with a wide range of approaches proposed for extracting structured knowledge from unstructured data sources \cite{Ji2020ASO}. Early methods relied heavily on rule-based systems and hand-crafted features to identify entities and relations in text \cite{10.1145/336597.336644}. With the advent of deep learning, neural network-based approaches have become increasingly popular, enabling more flexible and scalable KG construction \cite{zhang-etal-2017-position}.

One prominent line of work focuses on using distant supervision to automatically generate training data for relation extraction \cite{mintz-etal-2009-distant}. These methods assume that if two entities are mentioned together in a sentence and also appear in a knowledge base as subject and object of a relation, then that sentence is likely to express the relation. While distant supervision has been shown to be effective at scale, it often suffers from noise and incomplete coverage.

Another important direction is the development of unsupervised and semi-supervised methods for KG construction \cite{45634}. These approaches aim to reduce the reliance on large amounts of labeled data by leveraging techniques such as bootstrapping, graph-based inference, and representation learning. However, they often struggle with consistency and quality control issues.
More recently, there has been growing interest in using large language models for KG construction \cite{bosselut-etal-2019-comet}, \cite{chen-etal-2022-adaprompt}, \cite{yu_autokg_2021}. These methods take advantage of the vast knowledge captured in pretrained Language Models (LM) to generate KG triples through prompt engineering and fine-tuning. While promising, these LM-based approaches only produces triplets without canocalization, which makes portability and interoperability difficult. Additionally, some methods rely on vector-based similarity measures to deduce relationships between entities in KG, which yields good performance but falls short in interpretability \cite{chen_autokg_2023}.


As mentiond in Introduction, despite the significant progress in KG construction and LLM applications, performance, interpretability, coverage of proprietary documents, and interaction with other knowledge base remain issues. Our pipeline address these by grouding KG generation on ontology based on Wikidata schema, which ensures that output KG is human-readable and makes integrating with Wikidata or other KG easier; In the experiments below we show that these benefits can be also achieved on private documents with decent performance.

%% file: src/metho.tex
Our proposed approach for ontology-grounded KG construction using LLMs consists of four main stages: 1) Competency Question Generation, 2) Relation Extraction and Ontology Matching, 3) Ontology Formatting, and 4) KG Construction. Figure 1 provides an overview of the pipeline.

\subsection{Competency Question (CQ)-Answer Generation}
\textbf{The first step} in our pipeline is to generate a set of competency questions (CQs) and answers that capture the key information needs of the target domain. We employ an LLM to generate CQs based on the input documents. The LLM is provided with a set of instructions and examples to guide the generation process, encouraging the creation of well-formed, relevant questions that can be answered using the given documents. This step helps to scope the KG construction task within the knowledge domain, and ensure that the resulting KG aligns with the intended use cases. This also allows further ontology expansion by incorporating user-submitted domain-defining questions when interacting with the knowledge base, which serves as a user friendly interface of refining ontology by submitting new CQs and use our proposed pipeline to attach the incremental knowledge scope to existing ontology.

\subsection{Relation Extraction and Ontology Matching}
\label{sec:om}

During our preliminary experiments of prompting LLMs to directly generate ontology on documents, we noted that the LLM spontaneously recalled Wikidata knowledge in response, consistent with previous works \cite{semnani_wikichat_2023}. In our preliminary experiments, this behaviour also transfers to small 7B/14B models.

Following this direction, in \textbf{the second step} we extract relations from CQs and match them against Wikidata properties to better elicit model memories on Wikidata when constructing and using ontology. We first prompt LLMs to extract properties from CQ and write brief description on usage of extracted properties, including their domain and range, following editing guidelines of Wikidata. To match these properties against existing entries in Wikidata ontology, we pre-populate a candidate property list with all Wikidata properties after filtering out properties related to external database/knowledge base IDs. These extracted properties are then matched against the candidate list by a vector similarity search between description of properties. The representation for property is sentence embedding constructed from description of properties, and the top 1 closest candidate is retrieved for each extracted property. This matching result between each pair of extracted property and matched top 1 candidate is then vetted by LLM to see if they are really semantically similar as a final deduplication step. If a match is validated, the candidate property is added to the final property list; otherwise, the newly minted property is kept in the final list if we allow expansion from the candidate property list derived from Wikidata, and discarded when the final property list is required to be a subset of candidate property list. The first scenario is suitable for cases when no prior schema is known for the domain and some new properties outside of common ontology are expected, whereas the latter is for a known target list of possible properties.



\subsection{Ontology Formatting}
\textbf{In the third stage}, we use LLM to generate an OWL ontology based on the matched and newly created properties. We copy the description, domain and range field from all properties under Wikidata semantics. For new properties, LLM is prompted to infer and summarize classes for the domain and range of the relations to output a complete OWL ontology, following the format of copied Wikidata properties. This step ensures that the resulting KG is grounded in a formal, machine-readable ontology that captures relationships between entities, and close to the semantics of Wikidata for interoperability. 

\subsection{KG Construction}
\textbf{In the final stage}, we use the LLM to construct a KG based on the CQs and related answers grounded by the generated ontology in the previous stage. For each (CQ, answer) pair, LLM extracts relevant entities and maps them to the ontology using the defined properties. The output is a set of RDF triples that constitutes the final KG.

%% file: src/res.tex
\subsection{Experiment settings}
We evaluate our ontology-grounded approach to KG construction (KGC) on three datasets for KGC datasets: Wiki-NRE \cite{trisedya-etal-2019-neural}, SciERC \cite{luan2018multitask}, and WebNLG \cite{castro-ferreira-etal-2020-2020}. As Wiki-NRE and WebNLG are partially based on Wikidata and DBpedia (derived from Wikipedia contents), and in our proposed pipeline, Wikidata schema is utilized, we include SciERC for a more robust evaluation, since SciERC contains relation types that are not equivalent by nature to properties in Wikidata. 

We used a subset Wiki-NRE's test dataset containing 1,000 samples with 45 relation types following the split in \cite{zhang_extract_2024}, due to cost constraints. SciERC's test set contain 974 samples under a schema with 7 relation types. For WebNLG, we used test set in Semantic Parsing (SP) task, with 1,165 samples and 159 relation types. For evaluation, we adopt partial F1 on KG triplets based on standards in \cite{castro-ferreira-etal-2020-2020}. All experiments are conducted for one-pass 

We note in the previous reports that annotation in KGC reports may be incomplete in terms of both possible relation types and KG triplets \cite{zhang_extract_2024}, \cite{han2023information}. 

As our pipeline is designed to autonomously uncover knowledge structure with no prior assumption on knowledge schema, we report our result in two ways, corresponding to the two configurations of final de-duplication step in Section \ref{sec:om}:

\begin{enumerate}
    \item \textit{Target schema constrained}: In this setting, we match all relation types in test sets to its closest equivalent in Wikidata and constrict ontology to the relation universe in test set.
    \item \textit{No schema constraint}: In this setting, we do not filter matched ontology, even if they are not in schema of test dataset. This setting is close to real-life applications when processing documents with unknown schema.
\end{enumerate}

For property \textit{conjunction, evaluate for, compare, feature of} in SciERC, we select the closest properties proposed by LLM based on our subjective opinion. 

To highlight our system's competency, rather than directly prompting triplets, we parse output KG with RDF parser and extract all valid RDF triples for KG related to each document in test set, and present triplets to evaluation script for assessment. This ensures that our evaluation is on the generated KG ready to be consumed in downstream application. 

We test our pipeline on both Mistral-7B-instruct \cite{jiang2023mistral} and GPT-4o\footnote{https://openai.com/index/hello-gpt-4o/}. Due to cost constraints, we have only tested GPT-4o on \textit{target schema constrained} setting. For embedding property usage comment, we select bge-small-en \cite{bge_m3}. We use GenIE \cite{josifoski-etal-2022-genie}, PL-Marker \cite{ye-etal-2022-packed}, and ReGen \cite{dognin-etal-2021-regen} as fine-tuned baseline for Wiki-NRE, SciERC, and WebNLG dataset, respectively (collectively named \textbf{Non-LLM Baseline}). For LLM-based systems, we use results reported in \cite{zhang_extract_2024} for Wiki-NRE and WebNLG on the same Mistral model, and GPT-4 results in \cite{zhu_llms_2024} for SciERC. (collectively named \textbf{LLM Baseline}). We note that it is highly unlikely that Mistral-7B poses an advantage over an earlier version of GPT-4, when interpreting result of SciERC.

\subsection{Result}
\input{src/res_tbl}
Table \ref{tab:results} shows the performance of our method compared to state-of-the-art baselines on this subset. Our proposed approach exceeds all baseline under \textit{target schema constrained} setting on Wiki-NRE and SciERC datasets, while displaying a small performance regression when without schema constraint.  On WebNLG dataset, our pipeline maintained competitiveness against fine-tuned SOTA when constrained on target schema. These results validate the quality of KG generated by our pipeline, especially SciERC, whose semantics contains properties that are not native to Wikidata. We also note performance improvement when using GPT-4o.

\subsection{Discussion}

\subsubsection{Performance discrepancy on different grounding ontology}
It is worth noting that the relatively lower performance on \textit{no schema constraint} setting across all datasets is due to the fact that the LLM discovers a richer ontology than the predefined target schema. While this expanded schema may capture additional relevant information, it can hinder extraction performance when evaluated solely against the limited target schema. This showcases the trade-off between schema completeness and strict adherence to a predefined ontology, and our pipeline performs best on a large set of documents with a limited scope of knowledge, requiring a concise schema. 

Furthermore, the flipside of performance deficit in an absence of schema constraints, i.e. additional ontology entries outside of dataset-defined properties, cannot be evaluated against the dataset directly, as the ontology is not entirely covered by test set annotations. Hence, the virtue of \textit{no schema constraint} setting is to demonstrate that our pipeline can indeed provide a coverage of the properties in test set, though somewhat limited compared to baselines, when also capturing ontology outside test set schema, which is potentially more useful when discovering ontology on a novel document set with no expert knowledge in its schema conposition. This ability may be validated by manual evaluation on the full set of captured ontology in a future work down the line.

Nevertheless, the marginal performance deficit leaves room for improvement. Recent reports explored that long input context may pose challenge to LLMs even if such long context length is technically supported \cite{li2024longcontext}. We conjecture that aside from trimming grounding ontology, which hinders the knowledge coverage of our pipeline, few-shot fine-tuning on the new ontology or general pretraining in KG construction task may be helpful. We leave these as possible future directions.

\subsubsection{Utility of generated KG}

It should be emphasized that, while the selected evaluation tasks evaluate the correctness of extracted triplets, the extracted knowledge graph can do more than that. With ongoing discussion related to grounding LLM knowledge on trusted knowledge sources to reduce hallucination \cite{agrawal_can_2024}, explicitly generating KG provides a path to audit knowledge elicited when interacting with LLM, and with evidence demonstrating that LLM has the potential to reason on graph and generate an explicit path to retrieve required knowledge \cite{brei2024leveraging}, our pipeline may serve as a foundation for an interpretable QA system, where an LLM autonomously extracts ontology and deduces correct retrieval query based on the ontology when handling a set of unstructured document. The interpretability arise from the fact that KG and query could be understood and verified by users. Moreover, our usage of Wikidata schema offers potential interoperability with the whole Wikidata knowledge base, which safely expands the knowledge scope of QA system. We propose to continue research on this significant direction.

\subsubsection{Computational resources}

We note the growing concern of sustainability in LLM applications due to intensive requirement on computational resources. This pipeline consumes three separate LLM calls per document, plus one call per extracted relation. It is not straight forward to compare the carbon footprint of our approach compared to Non-LLM baselines, as our work at this stage does not require model fine-tuning, whereas all of the Non-LLM baselines employed various tuning techniques when producing the result. On the other hand, our smallest model adopted, Mistral-7B, is more than 10x larger in terms of parameter size compared to T5 models used in Non-LLM baselines. Larger models naturally require more powerful GPU clusters in terms of both GPU quantity and capability, but our zero-shot approach may provide an advantage in terms of resource cost compared to Non-LLM baselines when processing a small number of documents with no training requirement. 

When comparing with LLM baselines, we note that the approach by \cite{chen_autokg_2023}, \cite{zhu_llms_2024} consumes 1 and 2 LLM calls per document, respectively. However, we note that these baselines treat knowledge triplet as evaluation target, while we generate a formatted ontology at the end, which is more useful. Nevertheless, we recognize the performance burden and propose to explore techniques in fine-tuning and guided decoding to achieve better performance with smaller model and better reproducibility. 

%% file: src/res_tbl.tex
\begin{wraptable}{R}{0.55\textwidth}
    \centering
    \begin{tabular}{l|ccc}
        \hline
        Method & Wiki-NRE & SciERC & WebNLG \\
        \hline
        Non-LLM Baseline & 0.484 & 0.532 & \textbf{0.767} \\
        LLM Baseline & 0.647 & 0.07 & 0.728 \\
        Proposed (Mistral) & 0.66/0.60 & 0.73/0.58 & 0.74/0.68 \\
        Proposed (GPT-4o) & \textbf{0.71}/N/A & \textbf{0.77}/N/A & 0.76/N/A \\ 
        \hline
    \end{tabular}
    \caption{Partial F1 scores on test datasets. Best result is bolded. Results of proposed pipeline under two settings are presented as \textit{Target schema constrained/no schema constraint}.}
    \label{tab:results}
\end{wraptable}

%% file: src/con.tex
We have demonstrated the effectiveness of our ontology-grounded approach to KG construction using LLMs. By leveraging the structured knowledge in Wikidata, pretrained on LLM, and grounding KG construction with generated ontology, our pipeline is able to construct high-quality KGs across various domains while maintaining competitive performance with state-of-the-art baselines. Generated KGs that are conformant with Wikidata schema leaves possibly wide open, of building an interpretable QA system that has robust access to both common knowledge and proprietary knowledge base.

%% file: src/app.tex
\section{Sample generated KG}
This KG was generated under \textit{no schema constraint} setting for this document:
Mohammad Firouzi ( Born 1958 Tehran ) is a prolific Iranian musician , whose primary instrument is the barbat .
\begin{lstlisting}
<Prefixes and definition of dependencies omitted>
wd:Mohammad_Firouzi a wd:human ;
    rdfs:label "Mohammad Firouzi"@en ;
    wdt:occupation wd:Musician ;
    wdt:CountryOfCitizenship wd:Iran ;
    wdt:PlaceOfBirth wd:Tehran ;
    wdt:DateOfBirth "1958"^^xsd:date .
\end{lstlisting}
Note that in official annotation, only triplets related to place of birth and nationality exist, hence the evaluation will be penalized with low precision.

\section{Preprocessing of Wikidata schema}
To save space in LLM input context and mitigate performance drop on selected target schema when ontology is large, we only include commonly used properties by restricting data type on \textit{item, quantity, string, monolingual text, point in time.}\footnote{https://www.wikidata.org/wiki/Help:Data\_type}
To align with common pretraining objectives of LLM, we substitute entity identifiers (e.g. \textit{P19}) with its literal label (\textit{rdfs:label} in PascalCase (e.g. \textit{PlaceOfBirth}).

\section{Prompts}
All prompts are reused across all datasets.
\subsection{CQ generation}
We prompt LLM to generate up to 3 CQs per document for efficiency considering nature of test datasets, but note that this may be adjusted.
\begin{lstlisting}
Write competency questions based on the abstract level concepts in the document. Write questions that can be answered using the document only.
Write up to 3 questions per document. 
Below are the examples and follow the same format when generating competency questions: 

#### 
Document: Douglas Noel Adams (11 March 1952 - 11 May 2001) was an English author, humourist, and screenwriter, best known for The Hitchhiker's Guide to the Galaxy (HHGTTG). Originally a 1978 BBC radio comedy, The Hitchhiker's Guide to the Galaxy developed into a "trilogy" of five books that sold more than 15 million copies in his lifetime. It was further developed into a television series, several stage plays, comics, a video game, and a 2005 feature film. Adams's contribution to UK radio is commemorated in The Radio Academy's Hall of Fame.

####
Questions:
CQ1. What is the date of birth of Douglas Noel Adams?
CQ2. What is the date of death of Douglas Noel Adams?
CQ3. What is the occupation of Douglas Noel Adams?
CQ4. What is the country of citizenship of Douglas Noel Adams?
CQ5. What is the most notable work of Douglas Noel Adams?
CQ6. What is the original medium of The Hitchhiker's Guide to the Galaxy?
CQ7. In what year was The Hitchhiker's Guide to the Galaxy originally broadcast?
CQ8. How many books are in The Hitchhiker's Guide to the Galaxy "trilogy"?
CQ9. What other media adaptations were created based on The Hitchhiker's Guide to the Galaxy?

####
Document: 
{document to be processed}

####
Questions:
\end{lstlisting}
\subsection{CQ answering}
\begin{lstlisting}
Use the provided document to answer user query. If you don't know the answer, just say that you don't know, don't try to make up an answer.
Passage: {doc}
Query: {query}
\end{lstlisting}
\subsection{Relation extraction}
\begin{lstlisting}
You are an assistant in building a knowledge graph. Analyze the following competency questions and identify all relationships and concepts concepts mentioned in the question. 
Extract relation first, then describe the usage of each relation based on your understanding given the context of competency questions.
Afterwards, extract all relation-related concepts.
You should only extract properties between entities and literals, not entities themselves, or classes of entities. Therefore, not all CQs contain valid properties.
If you don't know the answer, just say that you don't know, don't try to make up an answer. 
Merge all relations into one list and all concepts into one list. 
Do not reply using a complete sentence, and only give the answer in the following format.

Below are the examples and follow the same format to extract the relations: 

#### 
Document: Douglas Noel Adams (11 March 1952 - 11 May 2001) was an English author, humourist, and screenwriter, best known for The Hitchhiker's Guide to the Galaxy (HHGTTG). Originally a 1978 BBC radio comedy, The Hitchhiker's Guide to the Galaxy developed into a "trilogy" of five books that sold more than 15 million copies in his lifetime. It was further developed into a television series, several stage plays, comics, a video game, and a 2005 feature film. Adams's contribution to UK radio is commemorated in The Radio Academy's Hall of Fame.

####
Questions:
CQ1. What is the date of birth of Douglas Noel Adams?
CQ2. What is the date of death of Douglas Noel Adams?
CQ3. What is the occupation of Douglas Noel Adams?
CQ4. What is the country of citizenship of Douglas Noel Adams?
CQ5. What is the most notable work of Douglas Noel Adams?
CQ6. What is the original medium of The Hitchhiker's Guide to the Galaxy?
CQ7. In what year was The Hitchhiker's Guide to the Galaxy originally broadcast?
CQ8. How many books are in The Hitchhiker's Guide to the Galaxy "trilogy"?
CQ9. What other media adaptations were created based on The Hitchhiker's Guide to the Galaxy?

####
Relations:
(date of birth, The date on which the subject was born.)
(date of death, The date on which the subject died.)
(occupation, The occupation of a person.)
(country of citizenship, The country of which the subject is a citizen.)
(notable work, The most notable work of a person.)
(genre, The genre or type of work.)
(publication date, The date or period when a work was first published or released.)
(has part, Indicates that the subject has a certain part, component, or element.)
(series, Indicates that the subject is part of a series, such as a book series, film series, or television series.) 

####
Document: 
{document to be processed}

####
Questions:
{CQs}

####
Relations:
\end{lstlisting}
\subsection{Ontology matching}
\begin{lstlisting}
Decide if the two properties are semantically similar in an ontology. 
You should say yes if you decide that these propties are similar, or if they are inverse properties.
Answer in "yes" or "no" only.
Property 1: {p1}
Property 2: {p2}
\end{lstlisting}
\subsection{Ontology formatting}
For properties under Wikidata schema, we retrieve \textit{schema:description, rdfs:domain, rdfs:range} for each property and include it in resulting ontology. Otherwise LLM is prompted to author ontology as so:
\begin{lstlisting}
Use the relations (properties) and their usage comments to build an ontology in RDF format.
If you don't know the answer, just say that you don't know, don't try to make up an answer.
Don't provide anything other than an ontology in RDF format.
Infer and summarize classes for domain and range of the relations across the concepts provided, and add these classes to relations only if required for clousre of relations.
For each relation, add relevant ontology entry for it. 
Add rdfs:comment based on the usage comments.
Use wdt: namespace for all relations discovered. Use entities under these prefixes if necessary:
@prefix rdf: <http://www.w3.org/1999/02/22-rdf-syntax-ns#> .
@prefix xsd: <http://www.w3.org/2001/XMLSchema#> .
@prefix rdfs: <http://www.w3.org/2000/01/rdf-schema#> .
@prefix owl: <http://www.w3.org/2002/07/owl#> .
@prefix wikibase: <http://wikiba.se/ontology#> .
@prefix schema: <http://schema.org/> .
@prefix wd: <http://www.wikidata.org/entity/> .
@prefix wdt: <http://www.wikidata.org/prop/direct/> .
Use turtle syntax.

Below is an example:

####
Relations:
(results, results: results of a competition such as sports or elections)

####
Ontology:
wdt:Results a wikibase:Property ;
    schema:description "results of a competition such as sports or elections" ;
    rdfs:label "results" ;
    rdfs:domain wd:referendum, wd:competition, wd:party conference, wd:sporting event ;
    rdfs:range wd:electoral result, wd:voting result, wd:sport result, wd:race result .

####
Relations:
{relation}

####
Ontology:
\end{lstlisting}
\subsection{KG generation}
\begin{lstlisting}
Your task is to construct a knowledge graph based on the provided ontology. 
Focus on understanding relationships from the question answer pair and document,  
and extract related entities, then mapping them to the ontology using the properties defined in the ontology. 
Do not include new properties other than those in ontology. Only use those properties in the ontology.
Output in turtle format following the ontology provided. 
You should only include knowledge in question answer pairs and the document.
Do not make up answers.

Use this ontology based on Wikidata as the starting point:
{ont}

Below is an example:

####
Document:
Douglas Noel Adams (11 March 1952 - 11 May 2001) was an English author, humourist, and screenwriter, best known for The Hitchhiker's Guide to the Galaxy (HHGTTG). Originally a 1978 BBC radio comedy, The Hitchhiker's Guide to the Galaxy developed into a "trilogy" of five books that sold more than 15 million copies in his lifetime. It was further developed into a television series, several stage plays, comics, a video game, and a 2005 feature film. Adams's contribution to UK radio is commemorated in The Radio Academy's Hall of Fame.

####
Question answer pairs:
Q: What is Douglas Adams an instance of?
A: Douglas Adams is an instance of human.

Q: What is Douglas Adams' sex or gender?
A: Douglas Adams' sex or gender is male.

Q: Where was Douglas Adams born?
A: Douglas Adams was born in Cambridge.

Q: Where did Douglas Adams die?
A: Douglas Adams died in Santa Barbara, California.

Q: When was Douglas Adams born?
A: Douglas Adams was born on 1952-03-11.

Q: On what date did Douglas Adams die?
A: Douglas Adams died on 2001-05-11.

Q: What occupation did Douglas Adams have?
A: Douglas Adams was a writer, comedian, and dramatist.

Q: What languages did Douglas Adams speak, write, or sign?
A: Douglas Adams spoke, wrote, or signed English.

Q: Where was Douglas Adams educated?
A: Douglas Adams was educated at St John's College, Cambridge and Brentwood School, Essex.

Q: What institution is Douglas Adams an alumni of?
A: Douglas Adams is an alumni of St John's College.

Q: What are some notable works by Douglas Adams?
A: Some notable works by Douglas Adams include The Hitchhiker's Guide to the Galaxy and Dirk Gently's Holistic Detective Agency.

Q: Was Douglas Adams a member of any notable organizations?
A: Yes, Douglas Adams was a member of Monty Python and The Independent on Sunday.

Q: What award did Douglas Adams receive?
A: Douglas Adams received the Locus Award for Best Science Fiction Novel.

Q: What is the Commons Category for Douglas Adams?
A: The Commons Category for Douglas Adams is "Douglas Adams".

####
Ontology:
@prefix rdf: http://www.w3.org/1999/02/22-rdf-syntax-ns# .
@prefix rdfs: http://www.w3.org/2000/01/rdf-schema# .
@prefix wdt: http://www.wikidata.org/prop/direct/ .
@prefix wd: http://www.wikidata.org/entity/ .
@prefix xsd: http://www.w3.org/2001/XMLSchema# .
wd:Douglas_Adams rdfs:label "Douglas Adams"@en ;
    wdt:InstanceOf wd:human ;
    wdt:SexOrGender wd:male ;
    wdt:PlaceOfBirth wd:Cambridge ;
    wdt:PlaceOfDeath wd:Santa_Barbara_California ;
    wdt:DateOfBirth "1952-03-11"^^xsd:date ;
    wdt:DateOfDeath "2001-05-11"^^xsd:date ;
    wdt:Occupation wd:writer ;
    wdt:Occupation wd:comedian ;
    wdt:Occupation wd:dramatist ;
    wdt:LanguagesSpokenWrittenOrSigned wd:English ;
    wdt:EducatedAt wd:St_Johns_College_Cambridge ;
    wdt:EducatedAt wd:Brentwood_School_Essex ;
    wdt:AlumniOf wd:St_Johns_College ;
    wdt:NotableWork wd:The_Hitchhikers_Guide_to_the_Galaxy ;
    wdt:NotableWork wd:Dirk_Gentlys_Holistic_Detective_Agency ;
    wdt:MemberOf wd:Monty_Python ;
    wdt:MemberOfOrganization wd:The_Independent_on_Sunday ;
    wdt:Award wd:Locus_Award_for_Best_Science_Fiction_Novel ;
    wdt:CommonsCategory "Douglas Adams"@en .
    
####
Document:
{doc}

####
Questions and Answer pairs:
{qa}

####
Ontology:
\end{lstlisting}